\pdfoutput=1

\documentclass[11pt]{article}

\usepackage{ACL2023}

\usepackage{times}
\usepackage{latexsym}
\usepackage{graphicx}

\usepackage[T1]{fontenc}

\usepackage[utf8]{inputenc}

\usepackage{microtype}

\usepackage{inconsolata}

\usepackage{multirow}
\usepackage{booktabs}

\title{Romanian Multiword Expression Detection Using Multilingual Adversarial Training and Lateral Inhibition}

\author{Andrei-Marius Avram$^{1}$, Verginica Barbu Mititelu$^{2}$, Dumitru-Clementin Cercel$^{1}$\\
$^{1}$University Politehnica of Bucharest, Faculty of Automatic Control and Computers\\
$^{2}$Romanian Academy Research Institute for Artificial Intelligence\\
\tt andrei\_marius.avram@stud.acs.upb.ro, vergi@racai.ro, dumitru.cercel@upb.ro\\
}

\begin{document}
\maketitle
\begin{abstract}

Multiword expressions are a key ingredient for developing large-scale and linguistically sound natural language processing technology. This paper describes our improvements in automatically identifying Romanian multiword expressions on the corpus released for the PARSEME v1.2 shared task. Our approach assumes a multilingual perspective based on the recently introduced lateral inhibition layer and adversarial training to boost the performance of the employed multilingual language models. With the help of these two methods, we improve the F1-score of XLM-RoBERTa by approximately 2.7\% on unseen multiword expressions, the main task of the PARSEME 1.2 edition. In addition, our results can be considered SOTA performance, as they outperform the previous results on Romanian obtained by the participants in this competition.

\end{abstract}

\section{Introduction}

The correct identification and handling of multiword expressions (MWEs) are important for various natural language processing (NLP) applications, such as machine translation, text classification, or information retrieval. For example, in machine translation, if an MWE is not recognized as such and is literally translated rather than as an expression, the resulting translation either is confusing or has the wrong meaning \cite{zaninello2020multiword}. In text classification, MWEs recognition can provide important information about the topic or sentiment of a text \cite{catone2019automatic}, while in information retrieval, MWEs can clarify the meaning of a query and improve the accuracy of search results \cite{englmeier2021aspect}.


The PARSEME COST Action\footnote{\url{https://typo.uni-konstanz.de/parseme/.}} organized three editions \citet{corpus-1.0, corpus-1.1, corpus-1.2} of a shared task that aimed at improving the identification of verbal MWEs (VMWEs) in text. This work improves the results obtained in PARSEME 1.2 \cite{corpus-1.2} for the Romanian language. We investigate the advantages of using Romanian monolingual Transformer-based \cite{vaswani2017attention} language models together with merging all the datasets for each language presented at the competition in a single corpus and then fine-tuning several multilingual language models on it. Additionally, for the latter, we aim to enhance the overall system's performance by generating language-independent features, with the help of two techniques, namely the lateral inhibition layer \cite{pais2022racai} on top of the language models and adversarial training \cite{lowd2005adversarial} between languages. 

Our experiments show that by employing these two algorithms, the results of the cross-lingual robustly optimized BERT approach (XLM-RoBERTa) \cite{conneau2020unsupervised} improve by 2.7\% on unseen MWEs when trained on the combined dataset. Additionally, we report state-of-the-art (SOTA) results with the monolingual training of Romanian Bidirectional Encoder Representations from Transformer  (RoBERT) \cite{dumitrescu2020birth} in comparison with the results obtained at the PARSEME 1.2 edition, achieving an F1-score of 60.46\%, an improvement of over 20\%.

\section{Dataset}

The PARSEME multilingual corpus was annotated with several types of VMWEs, to serve as training and testing material for the shared task. The quality of the manual annotation was further enhanced by a semi-automatic way of ensuring annotation consistency. For edition 1.2, the corpus contained 14 languages: Basque, Chinese, French, German, Hebrew, Hindi, Irish, Italian, Modern Greek, Polish, Portuguese, Romanian, Swedish, and Turkish. 

The types of VMWEs (i.e.,  universal, quasi-universal, and language-specific types) annotated therein are described in the annotation guidelines\footnote{\url{https://parsemefr.lis-lab.fr/parseme-st-guidelines/1.2/.}}.
The types of VMWEs annotated for Romanian are as follows: VID (verbal idiom) like "fura somnul" (eng., "steal sleep-the", "fall asleep''), LVC.full (light verb construction with a semantically bleached verb) like "da
citire" (eng., "give reading", "read''), LVC.cause (light verb construction in which the verb has a causative meaning) like "da foc" (eng., "give fire", "put on fire''), and IRV (inherently reflexive verb) like "se gândi" (eng., "Refl.Cl. think", "think").

The whole corpus version 1.2  contains 5.5 million tokens with 68k VMWEs annotations, split into train, dev, and test sets, on the one hand for controlling the distribution of unseen VMWEs both in dev with respect to test and in test with respect to train+dev, and on the other hand in ensuring a sufficient number of unseen VMWEs in the test set for each language. 
The Romanian training corpus contains 195k tokens in which 1,218 VMWEs are annotated. The Romanian dev set contains 134,340 tokens and 818 annotated VMWEs; the Romanian test set includes 685,566 tokens and 4,135 annotated VMWEs. The frequency of occurrence of VMWEs in Romanian ranges from 8\% (for LVC.full) to 22\% (for LVC.cause), with an average of 12\%, thus being quite redundant \cite{barbu-mititelu-etal-2019-romanian}.

\section{System Description}

\subsection{Monolingual Training}

We experiment with four BERT-based models  (first two monolingual and last two multilingual) for MWE identification using only the Romanian part of the PARSEME 1.2 corpus, namely the RoBERT, the Distilled Romanian BERT (Distil-RoBERT) \cite{avram-etal-2022-distilling}, the multilingual BERT (M-BERT) \cite{kenton2019bert}, and the XLM-RoBERTa \cite{conneau2020unsupervised}. We follow the standard sequence tagging procedure described in the original BERT model and fine-tune the embeddings produced by the last layer for the input tokens to predict the corresponding MWE labels using a feed-forward layer.

\subsection{Multilingual Training}

Our second and principal line of work here combines all the training sets of the corpora. Therefore, we train the two multilingual language models on the resulting dataset and then evaluate the models on the Romanian test set of the PARSEME 1.2 shared task. In addition, we improve the performance of the system by forcing the embeddings of the respective language models to depend less on their source language and more on the semantic specificities of an MWE using a lateral inhibition layer and adversarial training.

\begin{figure*}
    \centering
    \includegraphics[width=0.90\textwidth]{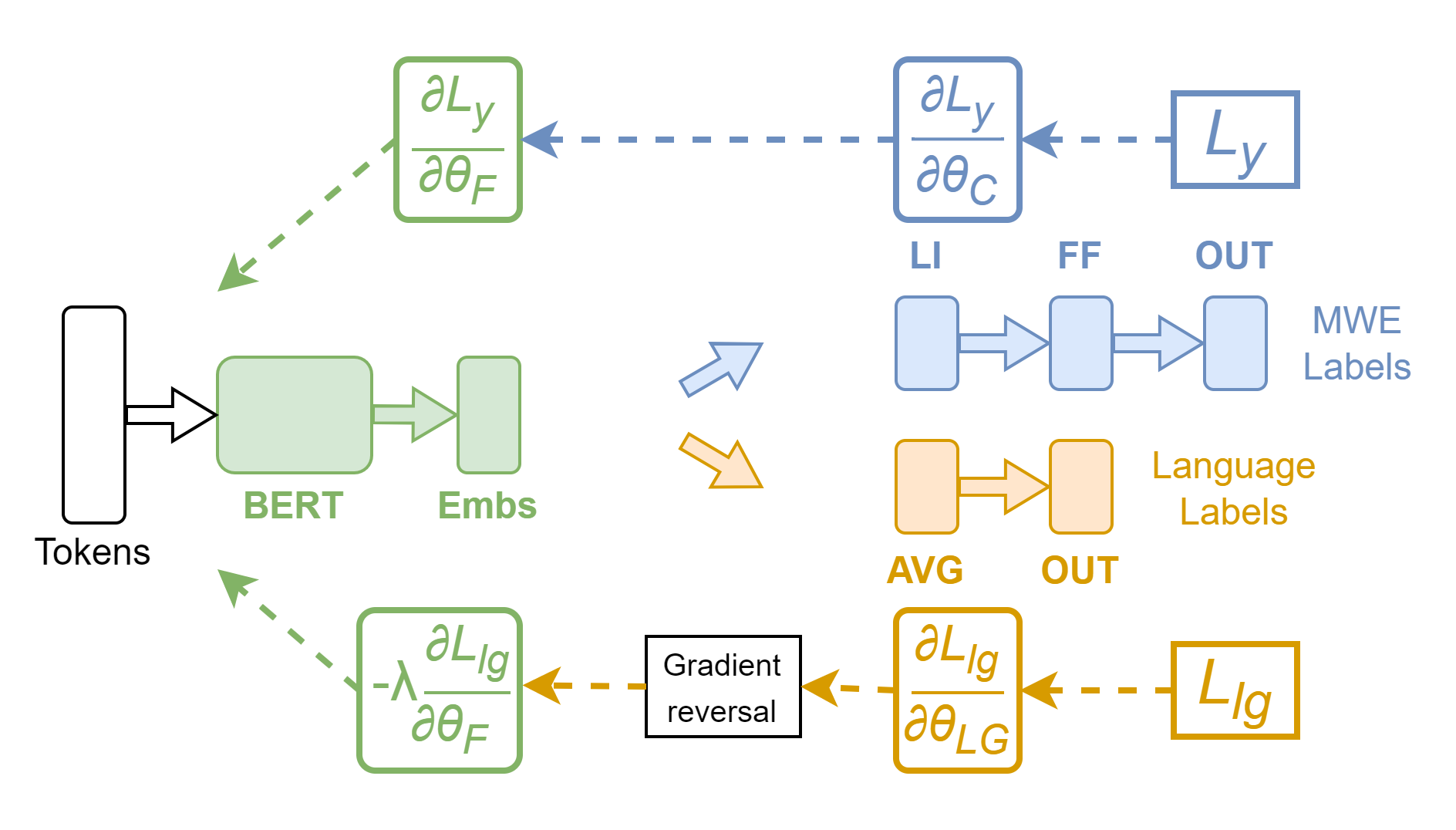}
    \caption{The multilingual training architecture. We use a multilingual BERT-based model to extract the embeddings from the input tokens (green). All these embeddings are fed into a classifier with a lateral inhibition layer to predict the MWE labels (blue) and into an adversarially trained language discriminator (orange). The block arrow depicts the forward pass, and the dotted arrow the backward pass.}
    \label{fig:sys_arch}
\end{figure*}

The general architecture of our multilingual training methodology is depicted in Figure \ref{fig:sys_arch}. It is divided into three major components: a multilingual BERT model that acts as a feature extractor $F$ and produces the embeddings of the tokens, a classifier $C$ whose role is to identify the MWEs in the given texts, and a language discriminator $LG$ whose role is to recognize the language of the input. We employ the lateral inhibition layer before feeding the embeddings to $C$ and adversarially train $LG$ by reversing its gradient before backpropagating through $F$. Further details on these two methods are given below.

\subsection{Lateral Inhibition} The neural inhibitory layer, modelled after the biological process of lateral inhibition in the brain, has been successfully used for the named entity recognition (NER) task in the past \cite{pais2022racai,avram2022racai,mitrofan2022improving}. We envisage that since the terms recognised by NER are just a subset of the MWEs identification, both being grounded in sequence tagging, introducing this layer into our model would also bring improvements in the final performance of our system. However, in the previous work, the neural inhibitory layer was mainly used to enhance the quality of the extracted named entities. In contrast, in this work, we employ it to achieve language-independent embeddings out of the multilingual transformer models.

The main idea behind the lateral inhibitory layer is quite simple. Given the embeddings $X$ produced by a language model and a weight matrix $W$ with a bias $b$, the output $Y$ of this layer is described in the following formula:

\begin{equation}
\label{eq:matrices}
Y = X * Diag(H(X * ZeroDiag(W^T) + b))
\end{equation}
where $Diag$ is a function that creates a matrix whose main diagonal is the vector given as input, $ZeroDiag$ is a function that sets a given matrix with the zero value on the main diagonal, and $H$ is the Heaviside step function.

Equation \ref{eq:matrices} works well for the forward pass. However, since the Heaviside step function is not differentiable, the lateral inhibition layer approximates the respective gradients with the gradients of the parameterized Sigmoid function \cite{Wunderlich2021}, a technique known as surrogate gradient learning \cite{8891809}. 

\subsection{Adversarial Training} Adversarial training of neural networks has been a highly influential area of research in recent years, particularly in fields such as computer vision with generative unsupervised models \cite{gui2021review}. Adversarial training has also been used to train predictive models \cite{zhao2022adversarial}, and in recent research, both multilingual and cross-lingual adversarial neural networks were introduced \cite{hu2019adversarial,guzman2022cross}. These networks are designed to learn discriminative representations that are invariant to language. In this study, we utilize the same methodology to learn task-specific representations in a multilingual setting, trying to improve the predictive capabilities of the employed multilingual transformer models.

Our methodology closely follows the domain adversarial neural network algorithm (DANN) \cite{ganin2016domain}, the difference here being that instead of reversing the gradient to create domain-independent features, we reverse it to generate language-independent embeddings out of the multilingual transformer models. As is the case for our system, DANN has in its composition a feature extractor $F$, a label classifier $C$, and a domain classifier $D$ that is replaced in our work with a language classifier $LG$. Thus, the gradient computation of each component can be formalized in the following equations: 

\begin{equation}
    \begin{array}{c}
        \theta_C = \theta_C - \alpha\frac{\partial L_y}{\partial \theta_C} \\
        \theta_{LG} = \theta_{LG} - \alpha\frac{\partial L_{lg}}{\partial \theta_{LG}} \\
        \theta_F = \theta_F - \alpha (\frac{\partial L_y}{\partial \theta_F} - \lambda \frac{\partial L_{lg}}{\partial \theta_F})
    \end{array}
\end{equation}
where $\theta_C$ are the parameters of the label classifier, $L_y$ is the loss obtained by the label classifier when predicting the class labels $y$, $\theta_{LG}$ are the parameters of the language classifier, $L_{lg}$ is the loss obtained by the language classifier when predicting the language labels $d$, $\theta_F$ are the parameters of the feature extractor, $\lambda$ is the hyperparameter used to reverse the gradients, and $\alpha$ is the learning rate.

\section{Results}

\subsection{Monolingual Training}

\begin{table*}
    \centering
        \begin{tabular}{|l|ccc|ccc|}
            \toprule
            \multirow{2}{*}{\textbf{Model}} & \multicolumn{3}{c|}{\textbf{Global MWE}} & \multicolumn{3}{c|}{\textbf{Unseen MWE}} \\
            & \textbf{P} & \textbf{R} & \textbf{F}1 & \textbf{P} & \textbf{R} & \textbf{F1} \\
            \midrule
            MTLB-STRUCT & 89.88 & 91.05 & 90.46 & 28.84 & 41.47 & 34.02 \\
            TRAVIS-mono & \textbf{90.80} & 91.39 & 91.09 & 33.05 & 51.51 & 40.26 \\
            \midrule
            RoBERT & 90.73 & \textbf{93.74} & \textbf{92.21} & \textbf{52.97} & \textbf{70.69} & \textbf{60.56} \\
            Distil-RoBERT & 87.56 & 90.40 & 88.96 & 41.06 & 62.77 & 49.65 \\
            M-BERT & 90.39 & 90.11 & 90.25 & 46.82 & 51.09 & 48.86 \\
            XLM-RoBERTa & 90.72 & 91.46 & 91.09 & 51.54 & 62.77 & 56.61 \\
             \bottomrule
        \end{tabular}
    \caption{The results of the models trained on the monolingual Romanian set.}
    \label{tab:mono}
\end{table*}

\begin{table*}
    \centering
        \begin{tabular}{|l|ccc|ccc|}
            \toprule
            \multirow{2}{*}{\textbf{Model}} & \multicolumn{3}{c|}{\textbf{Global MWE}} & \multicolumn{3}{c|}{\textbf{Unseen MWE}} \\
            & \textbf{P} & \textbf{R} & \textbf{F}1 & \textbf{P} & \textbf{R} & \textbf{F1} \\
            \midrule
            
            M-BERT & \textbf{91.34} & 88.46 & \textbf{89.88} & \textbf{49.90} & 48.12 & 48.99 \\
            M-BERT + LI & 90.78 & 88.85 & 89.81 & 45.06 & 45.15 & 45.10 \\
            M-BERT + Adv & 89.14 & \textbf{90.13} & 89.63 & 46.27 & \textbf{56.44} & \textbf{50.85} \\
            M-BERT + LI + Adv & 89.95 & 88.78 & 89.36 & 45.44 & 50.30 & 47.74 \\

            \midrule

            XLM-RoBERTa & \textbf{91.23} & 92.53 & \textbf{91.87} & 52.92 & \textbf{64.55} & 58.16 \\
            XLM-RoBERTa + LI & 91.12 & 92.02 & 91.02 & 52.11 & 61.19 & 56.28 \\
            XLM-RoBERTa + Adv & 89.45 & \textbf{92.87} & 91.12 & 54.91 & 63.96 & 59.09 \\
            XLM-RoBERTa + Adv + LI & 90.49 & 92.61 & 91.53 & \textbf{55.01} & 64.47 & \textbf{59.36} \\
            
             \bottomrule
        \end{tabular}
    \caption{The results of the multilingual models trained on the multilingual combined dataset and evaluated on the Romanian set. LI means lateral inhibition, and Adv means multilingual adversarial training.}
    \label{tab:multi}
\end{table*}

Table \ref{tab:mono} shows the results of our monolingual training.
We report both the overall scores (called global MWE) and the scores of the identified MWEs that do not appear in the training set (called unseen MWE), as well as the results of the best overall system (MTLB-STRUCT) \cite{taslimipoor2020mtlb} and the results of the best system on Romanian (TRAVIS-mono) \cite{kurfali2020travis}. All our monolingual models outperform  the MTLB-STRUCT and TRAVIS-mono systems by more than 8\% on unseen MWE, with RoBERT achieving an improvement of more than 20\%. We believe that this is due to the more intensive hyperparameter search that we performed and the text preprocessing which consisted of things like replacing the letters with diacritics in Romanian to the standard used in pre-training or making sure that the tokenizer produces cased subtokens\footnote{These text preprocessing techniques are suggested at \url{https://github.com/dumitrescustefan/Romanian-Transformers}.}. 

Both the highest global MWE and unseen MWE performance were achieved by the monolingual RoBERT model, with F1-scores of 92.21\% and 60.56\%, respectively. The second highest performance was obtained by the XLM-RoBERTa model, although it is a multilingual model. Thus, XLM-RoBERTa outperformed the other monolingual model, Distil-RoBERT, by 2.1\% on global MWE and 7\% on unseen MWE. This phenomenon has also been noticed by \citet{conneau2020unsupervised}, showing the raw power of multilingual models pre-trained on a large amount of textual data.

\subsection{Multilingual Training}

Table \ref{tab:multi} shows the results for the multilingual training of both M-BERT and XLM-RoBERTa.
As in the monolingual training case, XLM-RoBERTa achieves better performance, coming out on top with an F1-score of 58.16\% in comparison with the 48.99\% F1-score obtained by M-BERT. We also notice that the simple multilingual training (i.e., without lateral inhibition and adversarial training) improves the results of the two models when trained on the monolingual Romanian set.

The adversarial training improves the performance of both M-BERT and XLM-RoBERTa in multilingual training. At the same time, the lateral inhibition layer brought improvements only to the later when it was combined with adversarial training. Thus, by merging the two methodologies, we outperform the XLM-RoBERTa's results trained on monolingual data (i.e., around 2.7\% on unseen MWEs), which was the main target of the competition, being behind RoBERT with only 1.2\%. 

\section{Conclusions}

The detection and processing of MWEs play an important role in various areas of NLP. This paper made notable improvements in unseen Romanian MWE identification by employing a lateral inhibition layer and adversarial training to multilingual large language models like XLM-RoBERTa. This way, we were able to improve the results of XLM-RoBERTa. In addition, we achieved SOTA results on this task with a simple fine-tuning of RoBERT that involved a better hyperparameter search and text preprocessing pipeline, respectively. 

Future work considers an analysis of the language-independent embeddings produced in the multilingual training, together with more experiments on other languages, to validate the generalization of this approach. In addition, we intend to add these results in LiRo - the public benchmark for Romanian NLP models \cite{dumitrescu2021liro}.

\bibliography{anthology,custom}
\bibliographystyle{acl_natbib}

\end{document}